# A Model of Plant Identification System Using GLCM, Lacunarity And Shen Features

Abdul Kadir

Department of Electronic & Computer Engineering Technology, Faculty of Engineering Technology, Technical University of Malaysia Malacca, Malaysia

**ABSTRACT**

Recently, many approaches have been introduced by several researchers to identify plants. Now, applications of texture, shape, color and vein features are common practices. However, there are many possibilities of methods can be developed to improve the performance of such identification systems. Therefore, several experiments had been conducted in this research. As a result, a new novel approach by using combination of Gray-Level Co-occurrence Matrix, lacunarity and Shen features and a Bayesian classifier gives a better result compared to other plant identification systems. For comparison, this research used two kinds of several datasets that were usually used for testing the performance of each plant identification system. The results show that the system gives an accuracy rate of 97.19% when using the Flavia dataset and 95.00% when using the Foliage dataset and outperforms other approaches.

**Keywords: GLCM, Lacunarity, Plant identification system, PFT, Shen features**

*Corresponding author

Email: akadir64@gmail.com

# INTRODUCTION

Several researches have been proposed various kinds of methods in recognizing plants using a leaf during the last six years. Use of Probability neural network (PNN) was introduced by Wu, et al. [1] and it was also used by Kadir, et al. [2] and Prasvita and Herdiyani [3]. Zulkifli [4] proposed General Regression Neural Network for the leaf identification system. Radial basis probabilistic neural network (RBPNN) and Pseudo-Zernike moments were researched by Kulkarni, et al. [5]. Novotny and Suk [6] did a research in recognizing woody plant using combination of Fourier descriptor and a simple NN classifier. Support Vector Machine was involved in Sing, et al. [7]. Boman [8] investigated a kernel based import vector machine (IVM). Valliamal and Geethalakshmi [9] combined Genetic Algorithm (GA) and Kernel Based Principle Component Analysis (KPCA). Legendre moments in classifying Bamboo species were used by Singh, et al. [10]. Legendre Moments, Fourier descriptors, Zernike Moments and Chebyshev moments were tested by Suk, et al. [11].

To reach a highest performance of the plant identification systems, features are extracted from the shape, the color, the texture, the vein and also the edge of leaves. Use of features such as roundness, eccentricity and dispersion derived from the leaves' shape is common now [2][9]. Shape features are also generated from any kinds of moments or descriptors, such as Zernike moments, Pseudo-Zernike moments, Legendre moments and Fourier descriptors. Color moments that involves mean, standard deviation, skewness and kurtosis are also often used in identifying plants, such as in [2][9]. Other possibility is using color histogram [12]. Incorporating vein of leaves as features by using simple morphology operations has been introduced by Wu, et al. [1]. They used four features derived from leaf's vein. Texture features were also added in the leaf identification systems. Du, et al. [13] developed a plant recognition system based on fractal dimension features. Sumanthi and Kumar [14] combined filter Gabor and other features for plant leaf classification. Gray-Level Co-occurrence Matrix (GLCM) for identifying plants was used in several researches, such as in [5][15][16]. Fiel and Sablatnig [17] used SIFT features to describe the texture of the region. Local Binary Patterns (LBP) was applied by Ren, et al. [18] for leaf image recognition. Meanwhile, Lin, et al. [19] proposed Gabor transform and LBP operator for classifying plant leaves.

In this research, a simple Bayesian classifier was applied for identifying leaves. This classifier used several features that were derived from the shape, the color, the texture and the vein of leaves. A group of shape features extracted from Polar Fourier Transform [2] were included. Besides, two features called as Shen features were added. The texture features were generated from GLCM and lacunarity, that was described by Petrou and Sevilla [20]. The color features consist of mean, standard deviation, skewness and kurtosis of leaf colors. The vein features come from Wu features [1].

To ease in comparing this research to other researches, two common datasets were used. Fortunately, several researchers utilized a common dataset such as the Flavia dataset, that contains 32 kinds of plants. So, comparison among the performance of plant identification systems is enabled. The second dataset used in this research is the Foliage dataset, that contains 60 kinds of plants. For example, the Flavia dataset was used by [1][5][7][21] and the Foliage dataset was used in [2]. Using such datasets can help other researchers to judge the performance of their systems. Therefore, this research tried to use both of the datasets in order to compare these results and other ones.

**MATERIALS AND METHODS**

**Shape Features**

Shape features can be divided into three groups. First, the features are generated by Polar Fourier Transform. Second, the features are derived from the central moment. Third, two features are related to convex hull.

Polar Fourier Transform [22] is defined as

$$PF2(\rho,\phi) = \sum_r \sum_i f(\rho,\phi_i) \exp[j2\pi(\frac{r}{R}\rho + \frac{2\pi}{T}\phi)] \qquad (1)$$

where $0 \leq r < R$ dan $\theta_I = I(2\pi/T)$ $(0 \leq I < T)$; $0 \leq \rho < R$, $0 \leq \phi < T$, R is radial frequency resolution and T is angular frequency resolution.

The calculation of PFT is based on an image I = {f(x, y); 0≤x<M, 0≤y<N}. The image is converted from the Cartesian space to the polar space $I_p$ = { f(r,θ); 0 ≤ r < R, 0 ≤ θ < 2π }, where R is the maximum radius from the centre of the shape. The origin of the polar space becomes the centre of the space to get translation invariant. Then, the centroid $(x_c, y_c)$ is calculated by using formula

$$x_c = \frac{1}{M}\sum_{x=0}^{M-1} x, \quad y_c = \frac{1}{N}\sum_{x=0}^{N-1} y, \qquad (2)$$

and (r, θ) is computed by using:

$$r = \sqrt{(x-x_c)^2 + (y-y_c)^2}, \quad \theta = \arctan\frac{y-y_c}{x-x_c} \qquad (3)$$

The rotation invariance is obtained by ignoring the phase information in the coefficients. As a result, only the magnitudes of the coefficients are retained. To get the scale invariance, the first magnitude value is normalized by the area of the circle and all the magnitude values are normalized by the magnitude of the first coefficient. Then, the Fourier descriptors are computed as follows:

$$FDs = \{\frac{PF(0,0)}{2\pi r^2}, \frac{PF(0,1)}{PF(0,0)}, ...., \frac{PF(0,n)}{PF(0,0)}, ...., \frac{PF(m,0)}{PF(0,0)}, ...., \frac{PF(m,n)}{PF(0,0)}\} \qquad (4)$$

In this case, m is the maximum number of the radial frequencies and n is the maximum number of the angular frequencies. For experiments, m = 4 and n = 6 were applied.

Features called solidity and convexity are related to a convex hull [23]. Solidity is a ratio between the area of the leaf and the area of its convex hull. Convexity is defined as ratio between the convex hull perimeter of the leaf and the perimeter of the leaf. The formulas for both features are as follows:

$$solidity = \frac{area\ of\ leaf}{area\ of\ convex} \tag{5}$$

$$convexity = \frac{convex\ perimeter}{perimeter} \tag{6}$$

Graham Scan algorithm [24] is used to calculate the convex hull.

Shen [25] introduced use of two kinds of moments identified by $F_2'$ and $F_3'$. The formulas are as follows:

$$F_2' = \frac{M_3^{1/3}}{m_1} = \frac{\left\{\frac{1}{N}\sum_{n=0}^{N-1}[d(n)-m_1]^3\right\}^{1/3}}{m_1} \tag{7}$$

$$F_3' = \frac{M_4^{1/4}}{m_1} = \frac{\left\{\frac{1}{N}\sum_{n=0}^{N-1}[d(n)-m_1]^4\right\}^{1/4}}{m_1} \tag{8}$$

where

$$m_1 = \frac{1}{N}\sum_{n=0}^{N-1}d(n) \tag{9}$$

$F_3'$ describes the roughness of a contour. Both $F_2'$ and $F_3'$ are invariant to translation, rotation and scaling. Rangayyan [25] added a feature called mf, a good indicator of shape roughness. It is defined as

$$mf = F_3' - F_1' \tag{10}$$

where

$$F_1' = \frac{M_2^{1/2}}{m_1} = \frac{\left\{\frac{1}{N}\sum_{n=0}^{N-1}[d(n)-m_1]^2\right\}^{1/2}}{m_1} \tag{11}$$

**Color Features**

Color features are calculated by using mean (μ), standard deviation (σ), skewness (θ) and kurtosis (δ) calculations on a leaf. Here are the formulas:

$$\mu = \frac{1}{MN}\sum_{i=1}^{M}\sum_{j=1}^{N} P_{ij} \tag{12}$$

$$\sigma = \sqrt{\frac{1}{MN}\sum_{i=1}^{M}\sum_{j=1}^{N}(P_{ij}-\mu)^2} \tag{13}$$

$$\theta = \frac{\sum_{i=1}^{M}\sum_{j=1}^{N}(P_{ij}-\mu)^3}{MN\sigma^3} \tag{14}$$

$$\gamma = \frac{\sum_{i=1}^{M}\sum_{j=1}^{N}(P_{ij}-\mu)^4}{MN\sigma^4} - 3 \tag{15}$$

All calculations were done for R, G, B and gray components. Therefore, the total of color features is 16.

**Texture Features**

The texture features consist of features generated from GLCM and lacunarity. The features come from GLCM are very popular and has been used for several kinds of applications, including plants. Principally, GLCM arranges the neighbouring pixels in an image in four directions that are 135°, 90°, 45° and 0° [27]. The common distance between two pixels used in GLCM is one. Then, statistical computations are done by using several scalar quantities proposed by Haralick [28].

In this research, five measures from Haralick were used. The angular second moment (ASM) measures textural uniformity, the contrast measures coarse texture of the gray level, the inverse different moment (IDM) measures the local homogeneity a pixel pair, the entropy measures the degree of non-homogeneity of the texture and the correlation measures the linear dependency on the image. The formulas to calculate the fifth features are as follows:

$$ASM = \sum_{i=1}^{L}\sum_{j=1}^{L} GLCM(i,j) \tag{16}$$

$$Contrast = \sum_{i=1}^{L}\sum_{j=1}^{L}(i-j)^2 GLCM(i,j) \qquad (17)$$

$$IDM = \sum_{i=1}^{L}\sum_{j=1}^{L}\frac{GLCM(i,j)}{1+(i-j)} \qquad (18)$$

$$Entropy = \sum_{i=1}^{L}\sum_{j=1}^{L}GLCM(i,j)*\log(GLCM(i,j)) \qquad (19)$$

$$Correlation = \sum_{i=1}^{L}\sum_{j=1}^{L}\frac{(ij)GLCM(i,j)-\mu_i'\mu_j'}{\sigma_i'\sigma_i'} \qquad (20)$$

where

$$\mu_i' = \sum_{i=1}^{L}\sum_{j=1}^{L}i*GLCM(i,j) \qquad (21)$$

$$\mu_j' = \sum_{i=1}^{L}\sum_{j=1}^{L}j*GLCM(i,j) \qquad (22)$$

$$\sigma_i' = \sum_{i=1}^{L}\sum_{j=1}^{L}GLCM(i,j)(i-\mu_i')^2 \qquad (23)$$

$$\sigma_j' = \sum_{i=1}^{L}\sum_{j=1}^{L}GLCM(i,j)(j-\mu_j')^2 \qquad (24)$$

Lacunarity is a fractal measure was described by Petrou & Sevilla [20]. It can be used to differentiate between two fractals with the same fractal dimension. Features come from lacunarity are defined as follows:

$$L_s = \frac{\frac{1}{MN}\sum_{m=1}^{M}\sum_{n=1}^{N}P_{mn}^2}{\left(\frac{1}{MN}\sum_{k=1}^{M}\sum_{l=1}^{N}P_{kl}\right)^2} - 1 \qquad (25)$$

$$L_a = \frac{1}{MN}\sum_{m=1}^{M}\sum_{n=1}^{N}\left|\frac{P_{mn}}{\frac{1}{MN}\sum_{k=1}^{M}\sum_{l=1}^{N}P_{kl}} - 1\right| \qquad (26)$$

$$L_p = \left( \frac{1}{MN} \sum_{m=1}^{M} \sum_{n=1}^{N} \left( \frac{P_{mn}}{\frac{1}{MN} \sum_{k=1}^{M} \sum_{l=1}^{N} P_{kl}} - 1 \right)^p \right)^{1/p} \quad (27)$$

In this research, Ls, La, and Lp were used and applied to component R, G, B and the intensity in gray scale image as well. In this case, p = 2, 4 and 6 were investigated.

**Vein Features**

Vein features are features derived from vein of the leaf by using morphology operation, introduced by Wu [1]. There are four features, defined as follows:

$$V_1 = A_1/A, \quad V_2 = A_2/A, \quad V_3 = A_3/A, \quad V_4 = A_4/A \quad (28)$$

$A_1$, $A_2$, $A_3$ and $A_4$ are number of pixels that construct the vein and A is the area of the leaf. The vein of the leaf was constructed by using morphological operation called opening on the gray scale image with flat, disk-shaped structuring element of radius 1, 2, 3, and 4, respectively, and then subtracted the remaining image by the margin. The results are like a vein structures.

**The Plant Identification System**

Fig. 1 shows a block diagram of the identification system. At first, the leaf to be identified is preprocessed. Then, the leaf is segmented from its background. After that, features are extracted from the leaf. These features are processed by a Bayesian classifier by involving features from the references of leaves.

The classifier is based on Gaussian or normal probability density with equal covariance matrix. This classifier follows the rule of Bayes decision theory:

$$P(\omega_i | x) = \frac{p(x | \omega_i) P(\omega_i)}{p(x)} \quad (29)$$

where

$$p(x) = \sum_{i=1}^{c} p(x | \omega_i) P(\omega_i) \quad (30)$$

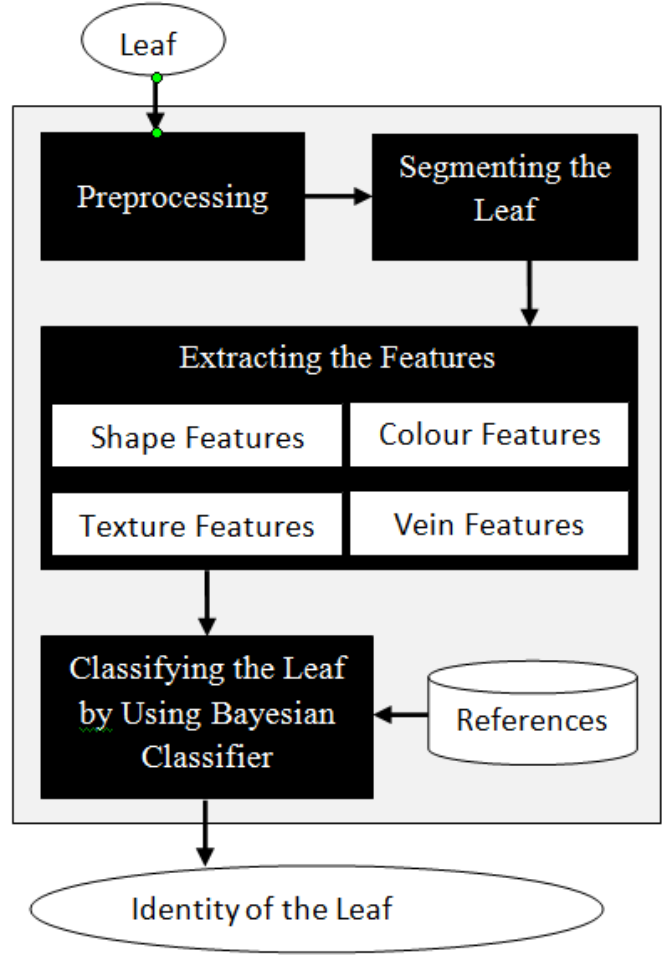

Fig. 1: Scheme of the plant identification system.

$P(\omega_i)$ is *a priori* probability of class $\omega_i$; $P(\omega_i | x)$ is *a posteriori* probability density function (pdf) of $x_i$; $p(x | \omega_i)$ is the class conditional pdf of x given $\omega_i$ where i = 1,2,…, c and c is number of classes. Then, x is assigned to the class $\omega_i$ if

$$P(\omega_i | x) > P(\omega_j | x), \forall j \neq i \tag{31}$$

In this case, the data in each class are distributed to the Gaussian distribution $N(m_i, S_i)$, where $m_i$ is the mean of the class $\omega_i$ and $S_i$ is the covariance matrix of the class $\omega_i$.

To measure the performance of the system, the accuracy of the system was calculated by using the following formula:

$$accuracy = \frac{n_r}{n_t} \tag{32}$$

In this case, $n_r$ is the relevant number of leaves and $n_t$ is the total number of tested leaves.

## RESULTS AND DISCUSSION

The Flavia and Foliage dataset were used to test the plant identification system. Data for testing and for references were separated. For the Flavia dataset, 10 leaves per species were used for testing purpose and 30 leaves per species were used as references. For the Foliage dataset, 20 leaves per species were used for testing purpose and 90 leaves per species were used as references.

The results of some experiments in combining various features for both datasets can be seen in Table 1. The table shows that combination of features came from PFT, solidity, convexity, color features, vein features, GLCM and lacunarity and Shen Features gave the optimum results either in the Flavia dataset or in the Foliage dataset.

Table 1: Experimental results

| No. | Features | Accuracy Rate | |
|---|---|---|---|
| | | Flavia | Foliage |
| 1 | PFT | 70.00% | 66.00% |
| 2 | PFT + Solidity + convexity | 76.80% | 69.75% |
| 3 | PFT + Solidity + convexity + color features | 92.50% | 91.33% |
| 4 | PFT + Solidity + convexity + color features + vein features | 94.37% | 92.08% |
| 5 | PFT + Solidity + convexity + color features + GLCM | 93.44% | 93.25% |
| 6 | PFT + Solidity + convexity + color features + vein features + GLCM | 94.69% | 94.17% |
| 7 | PFT + Solidity + convexity + color features + vein features + lacunarity | 95.94% | 93.00% |
| 8 | PFT + Solidity + convexity + color features + vein features + GLCM + lacunarity | 96.88% | 94.83% |
| 9 | PFT + Solidity + convexity + color features + vein features + GLCM + | 96.25% | 95.00% |

| No. | Features | Accuracy Rate | |
|---|---|---|---|
| | lacunarity + other shape features (eccentricity, roundness, and dispersion) | | |
| 10 | PFT + Solidity + convexity + color features + vein features + GLCM + lacunarity + Shen features | 97.19% | 95.00% |

Table 2 shows a comparison to other results that using the Flavia dataset. It can be seen that the plant identification using combination of features derived by GLCM, lacunarity and Shen features outperform other techniques. Based on the testing using the Foliage dataset, this approach also yields a better rate of accuracy than in [2] (95.00% vs. 93.08%).

Table 2: Comparison between this research's result and
the results of other methods using the Flavia dataset

| Method | Accuracy Rate |
|---|---|
| Fourier moment [7] | 46.30% |
| PNN [1] | 90.31% |
| PNN-PCNN [7] | 91.25% |
| LDA [21] | 94.30% |
| Pseudo Zernike Moment + RBPNN [5] | 94.52% |
| PFT + GLCM + PNN [2] | 94.69% |
| **Proposed system** | **97.19%** |

## CONCLUSIONS

A new approach in plant identification by using combination features generated from GLCM, lacunarity and Shen features has been implemented. This method gives a promising result. By using these features, features such as eccentricity, roundness, and dispersion to cope the shape of leaves can be eliminated. However, some experiments are still needed to combine GLCM and lacunarity and Shen features to other features and other classifiers.